\pdfoutput=1
\PassOptionsToPackage{table}{xcolor}
\documentclass[11pt]{article}

\usepackage{acl}
\usepackage{float}
\usepackage{placeins}
\usepackage{times}
\usepackage{latexsym}

\usepackage[T1]{fontenc}

\usepackage[utf8]{inputenc}

\usepackage{microtype}

\usepackage{inconsolata}

\usepackage{footnote}

\usepackage{booktabs}

\usepackage{amssymb}
\usepackage{graphicx}

\definecolor{burgundy}{HTML}{993F00}

\definecolor{teal}{HTML}{00A79D}
\definecolor{tomato}{HTML}{FF4600}
\definecolor{aqua}{HTML}{E0FFFB}

\usepackage{pifont} 

\usepackage{makecell}
\usepackage{ifthen}
\newboolean{showannotations}
\setboolean{showannotations}{false} 

\newcommand{\rubric}[1]{%
    \ifthenelse{\boolean{showannotations}}%
    {\textbf{\textcolor{red}{#1}}}%
    {}%
}

\newcommand{\system}{ScholaCite}

\title{Shallow Synthesis of Knowledge in GPT-Generated Texts: \\A Case Study in Automatic Related Work Composition}

\usepackage{authblk}

\author{
        Anna Martin-Boyle$^{\diamondsuit}$ \quad
        Aahan Tyagi$^{\diamondsuit}$ \quad
	Marti A. Hearst$^{\ddagger}$ \quad
	Dongyeop Kang$^{\diamondsuit}$ 
 \\
	$^\diamondsuit$University of Minnesota \quad $^\ddagger$UC Berkeley \\
	{\tt {\{mart5877,tyagi055,dongyeop\}@umn.edu} hearst@berkeley.edu} 
}

\begin{document}
\maketitle
\begin{abstract}

Numerous AI-assisted scholarly applications have been developed to aid different stages of the research process.
We present an analysis of AI-assisted scholarly writing generated with \system, a tool we built that is designed for organizing literature and composing Related Work sections for academic papers. Our evaluation method focuses on the analysis of citation graphs to assess the structural complexity and inter-connectedness of citations in texts and involves a three-way comparison between (1) original human-written texts, (2) purely GPT-generated texts, and (3) human-AI collaborative texts. We find that GPT-4 can generate reasonable coarse-grained citation groupings to support human users in brainstorming, but fails to perform detailed synthesis of related works without human intervention. We suggest that future writing assistant tools should not be used to draft text independently of the human author.

\end{abstract}

\section{Introduction}
The rapid acceleration of knowledge creation is a defining feature of modern research. \citet{Bornmann2021} report that between 1952 and 2020, the corpus of scientific literature has expanded at a growth rate of 5.08\%, doubling every 14 years. Some fields are growing even more rapidly; \citet{Krenn2023} estimate a doubling rate of almost two years for machine learning and artificial intelligence papers between 1994 and 2022. 

\begin{figure}[t!]
    \centering    
    \includegraphics[width=0.48\textwidth,trim={0 0 14cm 0},clip]{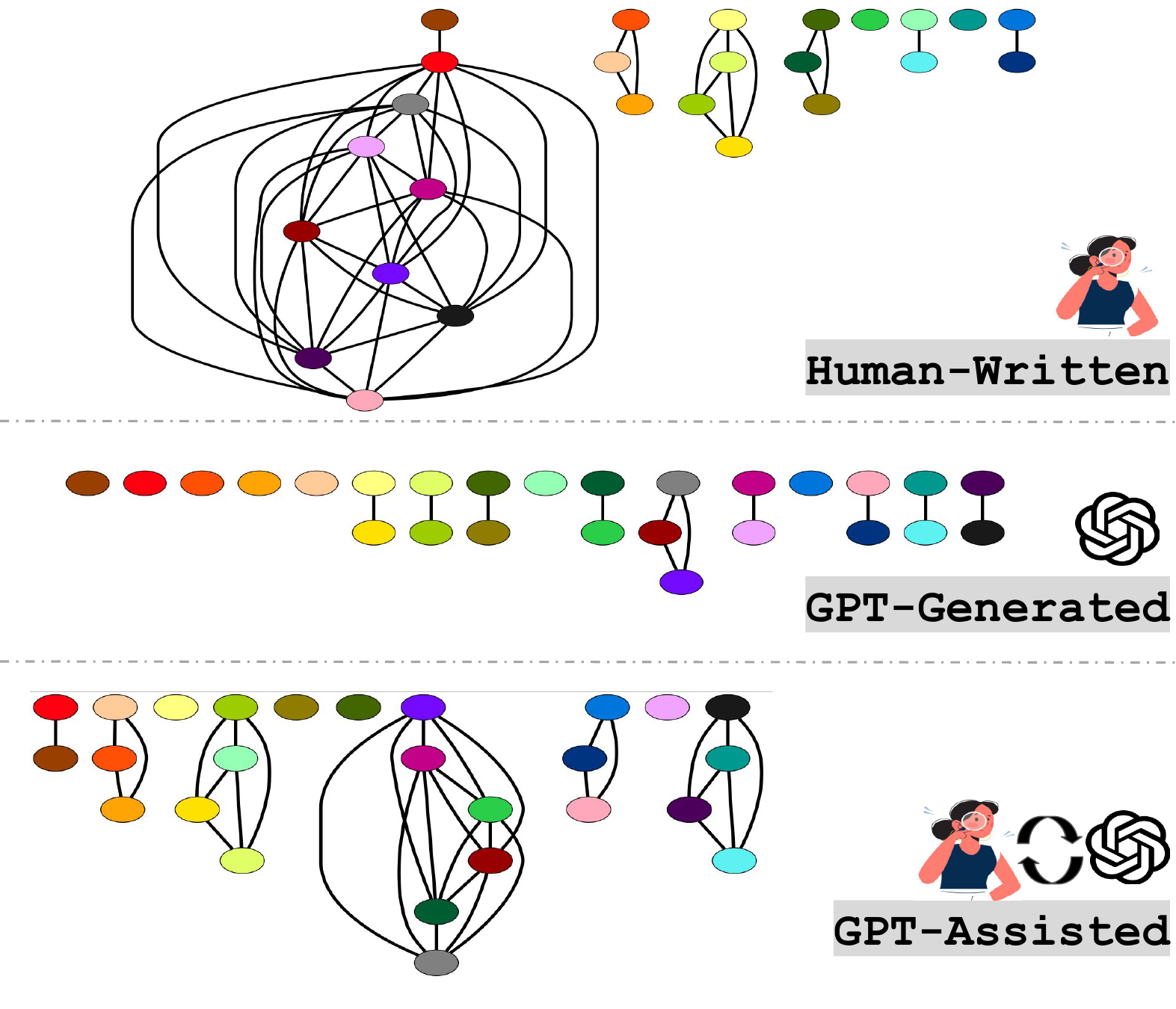}
    \caption{
    Citation graphs extracted from (top) human-written related work section of a scholarly paper, (middle) a GPT-assisted version composed using the ScholaCite application, and (bottom)a GPT-generated version.
    Nodes represent cited works and edges indicate concurrence of citations within the same sentence, where each citation is assigned a unique color that is consistent between the three samples (e.g., citation [1] is \textcolor{burgundy}{brown} in each illustration). This highlights the \textit{relative interconnectedness of the human-written texts compared to the GPT-generated text}.
    \label{fig:graphs}
    \vspace{-4mm}
    }
\end{figure}

In response to this growth rate, numerous AI-assisted scholarly applications have been developed to aid different stages of the research process, including literature discovery, reading interfaces, and writing tools. Scholars are also using ChatGPT \cite{10.5555/3495724.3495883} directly in the academic writing process. For example, \citet{Savelka2023Can} found 1,226 publications between 2022 and 2023 where the authors acknowledged using ChatGPT to write their paper. However, little work has been done to formally evaluate the outputs of these AI-assisted scholarly texts. It is important to study scholar-AI interactions because writing plays a key role in developing researchers' understanding \cite{doi:10.1177/1077800408318280}, not just disseminating results.

Our aim is to present an analysis comparing human-written, GPT-generated, and GPT-assisted related work sections with the goal of quantifying the ability of AI text to perform the synthesis and contextual framing that characterizes quality academic writing. This requires us to first define what makes a related work section ``high-quality'', and second, to develop a method for objectively measuring relative quality. For our purposes, we define quality using the following description from a guide on writing literature reviews \cite{cmu_litreview}\footnote{\url{https://www.cmu.edu/student-success/other-resources/handouts/comm-supp-pdfs/literature-reviews.pdf}}: 
\begin{quote}
    A literature review selects relevant past literature and \textbf{CONNECTS}, \textbf{SYNTHESIZES}, and sometimes \textbf{EVALUATES} these texts/studies, putting the authors in conversation with each other .
\end{quote}  
In addition to the above definition, the guide states that an author should establish their work's relevance to their research field, demonstrate their understanding of the preexisting discourse and findings on the topic, and explain the research gap that the new work fills. 
This definition is based on the \textit{author's choice of which works to discuss together} and how well those choices \textit{support the overall framing of their work}. 

We initially attempted a one-to-one comparison between the citation occurrences in the GPT-assisted text and the original human text. However, it became clear that this is not the right comparison  due to the open-ended nature of scholarly composition. For any given paper, there may be several valid ways to compose a related work section that best frames the work in progress. It is insufficient to sort citation sentences (citances) into True-Positive/False-Positive/False-Negative categories when the decision to use a citation is dependent on the larger context of the paper and the goals of the author. Furthermore, this analysis is difficult to automate and may involve some subjectivity. 

To address these evaluation challenges, we developed a novel evaluation method based on the \textbf{analysis of citation graphs}, where nodes are cited works and the edges represent concurrences of cited works within the same sentence (see Figure \ref{fig:graphs}). Examining the structural complexity and inter-connectedness in citation graphs is a useful proxy for how well  texts perform literary synthesis, via a method  that is objective, reproducible, and scalable. Furthermore, this method allows quantitative assessment of how effectively the text integrates and contextualizes multiple sources, a key component of high-quality related work sections. It offers a more objective lens through which to evaluate the ability of GPT-assisted compositions in achieving the essential goal of related work composition, which is not only to gather related works, but also to combine diverse pieces of research into a coherent narrative.

Our contributions are as follows:
\begin{itemize}
    \item We present \system, a GPT-assisted tool for organizing citations and composing related work sections;
    \item We introduce an evaluation method that compares GPT-generated (or GPT-assisted) text to human-written text using citation graph analysis; and
    \item We provide new insights into the capabilities and limitations of LLMs in emulating literature review composition, informing guidelines for appropriate AI usage to maintain scholarly rigor.
\end{itemize}

\section{Related Work}
\paragraph{Automatic Related Work Generation Approaches} 
There are numerous methods that have been proposed for automatically generating related work sections. 
\citet{AbuRaed2020Automatic} established a benchmark by introducing a neural sequence learning model that was able to produce citation sentences for related work sections. 

\citet{Zekun2021Automatic} introduced SERGE, A BERT-based system that uses sentence extraction and reordering, addressing previous issues of informativeness and citation delays.
\citet{Chen2021Capturing} developed a relation-aware related work generator, which generates abstracts for related work sections by considering the content dependencies between documents. In follow-up work, \citet{Chen2022Target} proposed an abstraction target-aware related work generator that used contrastive learning to better distinguish the target paper from referenced works. 
Focusing on the importance of topic-driven summarization, \citet{Cong2010Towards} created a methodology that prioritizes the motivations behind citations and the logical structure of related work sections, using hierarchical keyword arrangements to guide summarization, \citet{Wang2020ToC} introduced ToC-RWG, an unsupervised model that uses topic modeling and citation information, and \citet{Darsh2021Generating} utilized a content planning model to produce a tree structure of citations for subsequent lexicalization. 
In meta-analyses, \citet{Xiangci2022Automatic} underscore the need for standardization in problem formulation and evaluation, suggesting that future studies in related work composition could benefit from a more unified approach to these challenges, and  \citet{Justitia2022Automatic} looked at research trends and future directions in the field of related work summarization.
The approaches outlined in this subsection rely purely on language modeling and computational techniques for the automatic composition of related work sections, while we suggest a collaborative approach between human authors and AI assistants.
\paragraph{Automated Systematic Literature Review Applications} 
The task of performing systematic literature reviews has become increasingly complex with the exponential growth of available academic publications. This has led to significant interest in the development of automated tools and frameworks to guide researchers through the process. 
\citet{Tauchert2020Towards} explored the practical implications of automating the systematic literature review process, showing the potential of AI in extracting and summarizing key information.
\citet{Silva2021roadmap} also provided a roadmap for systematic literature review automation, focusing on the integration of various AI methodologies. They created a comprehensive guide for deploying techniques such as topic modeling, keyword extraction, and clustering to structure and synthesize literature.
\citet{Sahlab2022Knowledge} used knowledge graphs constructed from extracted data to streamline the literature review process, by creating visual representations of relationships and trends within the literature. 
Unlike these previous systems, we focus on the evaluation of the outputs of our AI-assisted related work generation system, applying citation graph analysis to understand the comparative quality of cited work synthesis in artificial texts.
\paragraph{The Use of ChatGPT in Academia} 
The emergence of Large Language Models (LLMs) like ChatGPT has already had an influence on academic and educational research methodologies. 
\citet{Karakose2023Utility} provides an exploration into the various applications of ChatGPT, emphasizing its role in enhancing creativity and productivity in academic research. Complementing Karakose's more optimistic perspective, \citet{Rahman2023ChatGPT} presents a more balanced point of view, pointing out critical issues regarding ChatGPT's accuracy, and recommending that reliance on its outputs without thorough verification may propagate inaccuracies in academic work. \citet{Buruk2023Academic} and \citet{AlZaabi2023ChatGPT} provide an ethical analysis, recommending transparency and the establishment of guidelines and ethical frameworks for the use of ChatGPT in academia, due to the risks of plagiarism and the lessening of academic integrity. 
The discourse around the use of LLMs like ChatGPT contains tension between the potential for such models to augment the research process, and the need for caution and guidelines to avoid the propagation of inaccuracies. 
We contribute to the discourse on the practical implications of using AI in scholarly writing by analyzing the ability of GPT-4 to synthesize cited works.

\section{Proposed Methods}
\begin{figure*}[ht]
    \centering
    \includegraphics[width=0.99\textwidth]{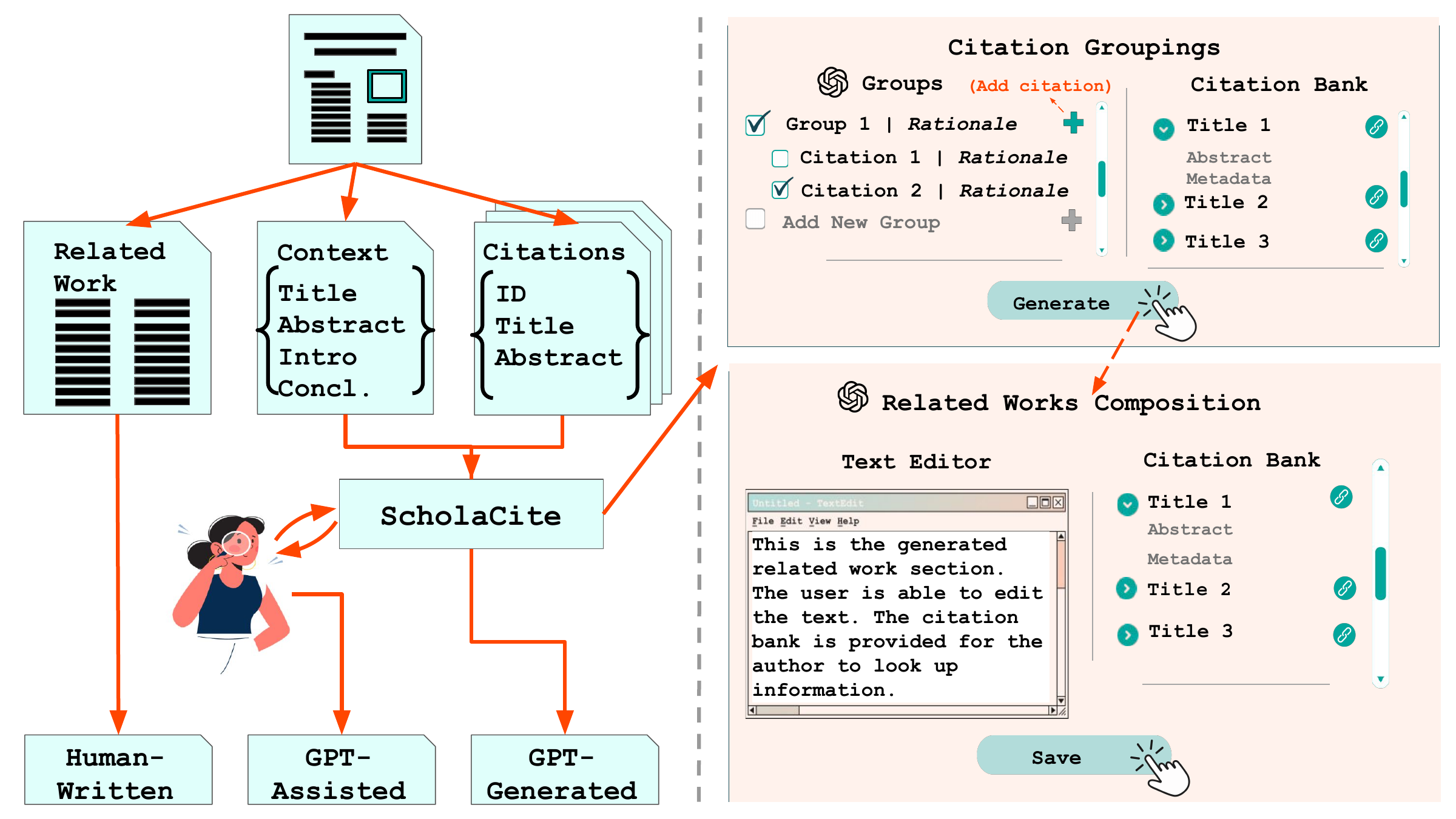}
    \caption{
    This workflow diagram shows the related work generation process for each condition: original human-written text, GPT-assisted text with ScholaCite, and GPT-generated text with ScholaCite.}
    \label{fig:workflow}
\end{figure*}

This section describes the dataset we assembled, our methods for related work generation, and the analytical techniques used in our comparative analysis of human-written, GPT-generated, and GPT-assisted related work sections. First, we introduce the task: creating related work sections for works in progress.  Next, we describe the ScholaCite system developed to generate texts under two different conditions (GPT-generated and GPT-assisted writing). Next, we present the dataset creation and preprocessing pipeline that produced the texts analyzed in this study. Finally, we provide an overview of the citation graph analysis metrics calculated to quantify and compare related works sections. See our workflow in Figure \ref{fig:workflow}.

Since we are studying GPT-generated texts, GPT-4 was used to generate our data. Therefore, some of the examples in this paper are synthetic, which is denoted with the phrases ``GPT-Assisted'' and ``GPT-4'' as in Table \ref{tab:samples}. AI-assistance was otherwise not used in our experiments or in the writing of this paper. 

\subsection{Task} 
To evaluate texts in a way that reflects the importance of author decision-making and full-paper contexts, we set up a test bed that consists of ten previously written papers, and  for each paper, the abstracts that the paper cites within its related work section. 
For evaluation, the LLM is provided with both the paper and its abstracts  with the original related work section redacted; we call this representation the ``work-in-progress'' or WIP, intended to simulate a paper that is partially written but lacking a related work section.
 For the evaluation task, we examine the following three conditions:
\begin{enumerate}
    \item The original related work section written by the human author(s) of the scholarly paper;
    \item A related work section written by a human interactively via  \system; and
    \item A related work section generated by GPT-4  without human intervention.
\end{enumerate}

\subsection{\system \ Application}
The \system \ web application was developed to facilitate the collaborative composition of related work sections in scholarly articles. It integrates GPT-4 to provide initial suggestions for citation groupings and rationales for each grouping. It also drafts text for the related work section based on those groupings. 
This two-step design first allows the user to organize and refine citation groupings, providing more coherent citation clusters for GPT-4 to use as an outline when drafting the text. In preliminary prompting experiments, we found that the two-step process yielded better-structured texts than simply providing the cited work abstracts and asking GPT-4 to draft a coherent related work section.

For the purposes of this study, the initial groupings and rationales were pre-generated using GPT-4 due to the time it takes for GPT-4 to generate groupings (approximately six minutes per sample). Authors can refine and edit these suggestions until they are satisfied with the organization and narrative of the related work section. When the author is done refining citation groupings, GPT-4 generates a related work section for the author to edit. See Figure \ref{fig:workflow} for a conceptual diagram of the application. 

\paragraph{GPT-4 Prompting}
For our experiments, we used the GPT-4 version that was accessible in October of 2023. This model had a context window size of 8,192 tokens and contained training data up to September 2021. Because of the short context-window size, GPT-4 was prompted in a zero-shot setting, except that examples were provided of how its output should be formatted in JSON. Following are descriptions of our prompting techniques for the generation of citation groups and the generation of related work sections. To read the prompts verbatim, see Appendix \ref{app:prompts}.

\paragraph{Step 1: Generation of Citation Groups}
First, we prompted GPT-4 to organize citations into coherent groups based on topic similarity and relevance to the work-in-progress paper. The prompts were provided in two stages because of the short context window. The objective of the first stage was to construct a grouping based only on the titles of related works, and the objective of the second stage was to refine the groupings one group at a time, given the abstracts of the related works. GPT-4 was also prompted to generate a rationale for why each group supports the work-in-progress, and for why each citation should belong in its group. Lastly, GPT-4 was prompted to format its output in a specified JSON format for easy post-processing.

\paragraph{Step 2: Generation of Related Work Section}
The ScholaCite system runs a related work generation script when the user requests it to draft the related work section based on their completed citation groupings. The groupings are provided in JSON format, and GPT-4 is asked to draft the related work section by generating a distinct paragraph for each group based on the group name, citations, and rationales. It is asked to produce a narrative that synthesizes the related works and frames the contribution of the work-in-progress. 

\subsection{Data}
We simulated scholarly works-in-progress (WIPs) using award-winning papers from the 2023 ACL conference\footnote{\url{https://2023.aclweb.org/program/best_papers/}}\footnote{the papers we used in our study are licensed under the \href{ https://creativecommons.org/licenses/by/4.0/ }{Creative Commons Attribution 4.0 International License}, and are permitted to be used for research purposes.}. The titles and citation references for each of the ten papers are included in Tables \ref{tab:graphs1} and \ref{tab:graphs2} in Appendix \ref{app:graphs}. These papers were selected due to being published after the cutoff date for GPT-4's training data at the time of our experiments to avoid contamination. They are all in English and, being written by scholars, follow the typical conventions of academic writing. The content of each paper falls within natural language processing domains. 
For clarity, we refer to our simulated works-in-progress as ``WIPs'', and the papers that are cited in the WIPs related work sections will be referred to as ``cited papers''. The total number of related work section citations varied dramatically between 8 and 52 citations. 
\subsubsection{Preprocessing} We preprocessed the documents with the help of Semantic Scholar's Academic Graph API\footnote{\url{https://www.semanticscholar.org/product/api}}. The data representation used in our study includes the title, abstract, introduction, related work, conclusion, and citations from the WIP's related work section. The title, abstract, introduction, and conclusion represent the WIP, while the related work section represents the original related work section. Each citation in the citation list includes its bibliographic metadata, including title, abstract, authors, year, and a URL for its Semantic Scholar paper page. Only the citations from the related work section are included in the citation list, since citation functions \citep{teufel-etal-2006-annotation} vary from section to section \citep{jurgens-etal-2018-measuring}. For example, citations may be used in the Introduction of the paper that would not play a meaningful role in the related work section. In-text citations in the abstract, introduction, and conclusion texts are masked with \textsc{citation}. 

 \begin{table}[t]
    \centering
    \begin{tabular}{p{.1\textwidth}p{.32\textwidth}}\toprule
        \textbf{Metric} & \textbf{Rationale} \\\midrule
        \# edges & Counts concurrences of citations within the same sentence. \\\midrule
        Avg. node degree & Measures the average number of concurrences per citation. A higher average degree suggests that many citations are discussed in conjunction with two or more other citations. \\\midrule
        Density & Ratio of the number of edges to the number of possible edges; measures the overall interconnectivity of the citation network by showing how close the network is to being a complete graph. \\\midrule
        Clustering coef. & Measures the degree to which citations tend to cluster together. High clustering indicates that certain groups of citations are frequently discussed together. \\\bottomrule
    \end{tabular}
    \caption{The graph metrics we selected for analysis and our rationale for selecting them.}
    \label{tab:metrics}
\end{table}

\subsection{Procedures}
The information in this section details the specific steps taken to produce texts for comparative analysis across the three conditions.
\paragraph{Condition 1}
The original related work section was extracted into a text file for comparison with GPT-assisted and GPT-generated sections.

\paragraph{Condition 2}
To prepare to write the GPT-assisted related work sections, one of the authors of this study read the title, abstract, introduction, and conclusion of each paper, and the titles and abstracts of each cited work. They then applied \system 
\ to each paper in the dataset, using the grouping page until they were satisfied with the quality of the groupings. The author then used the ScholaCite text editor to refine the generated related work section. This process took between 40 minutes and 2.5 hours per sample, based on how many citations were included. 

\begin{table*}[ht!]
    \centering
    \begin{tabular}{c|p{.17\columnwidth}p{.17\columnwidth}p{.17\columnwidth}|p{.25\columnwidth}p{.25\columnwidth}p{.25\columnwidth}}\toprule
        \textbf{Statistic} & \textbf{Human  (avg)} & \textbf{Assisted  (avg)} & \textbf{GPT-4  (avg)} & \textbf{Human vs. Assisted (p)} & \textbf{Human vs. GPT-4 (p)} & \textbf{Assisted vs GPT-4 (p)}\\\midrule
        \makecell{Number of \\edges} & 48.30 & 38.50 & 12.30 & \makecell{56.5 \\{\small{(1.0000)}}} & \makecell{84.5 \\{\small{(0.0202)}}} & \makecell{79.0 \\{\small{(0.0747)}}}\\\midrule
        \makecell{Average \\node degree} & 3.23 & 2.69 & 0.70 & \makecell{62.0 \\{\small{(1.0000)}}} & \makecell{93.0 \\{\small{(0.0020)}}} & \makecell{94.0 \\{\small{(0.0013)}}} \\\midrule
        Density & 0.14 & 0.13 & 0.02 & \makecell{61.0 \\{\small{(1.0000)}}} & \makecell{65.0 \\{\small{(0.0002)}}} & \makecell{99.0 \\{\small{(0.0003)}}} \\\midrule
        \makecell{Cluster \\coefficient} & 0.66 & 0.63 & 0.15 & \makecell{64.0 \\{\small{(0.9203)}}} & \makecell{91.0 \\{\small{(0.0034)}}} & \makecell{97.5 \\{\small{(0.0004)}}} \\\bottomrule
        
    \end{tabular}
    \caption{Citation graph analysis results 
    reveal significantly higher connectivity and clustering in human-written texts compared to GPT-generated texts, suggesting superior integration and synthesis of cited works. GPT-assisted texts exhibit intermediate performance, with metrics closer to human levels. There is no significant difference between the Mann-Whitney scores of the human-written and GPT-assisted texts.
    }
    \label{tab:results}
\end{table*}

\paragraph{Condition 3}
 We generated the GPT-4 samples by using the system without editing the groupings or the output related work section. In this way, the samples were composed following the same grouping and drafting sequence as for Condition 2, but without human intervention.

\subsection{Evaluation: Citation Graph Analysis}
To evaluate the outputs of \system, we conducted a comprehensive statistical analysis to compare the citation graphs of human-written versus GPT-generated texts, since analyzing citation patterns is a more objective and scalable way compared to manual inspection of the generated text. We constructed citation graphs where nodes represent individual citations and edges indicate the concurrence of two citations within the same sentence. This method aims to analyze and compare the structural complexity and inter-connectedness of citations in human-written and GPT-generated texts. For each paper, we calculated the number of edges, average node degree, density, and clustering coefficient, the intuition being that these metrics can serve as a proxy measure of the level of synthesis of cited works (for explanations of each of these metrics, see Table \ref{tab:metrics}). To test for the significance of the differences between the human-written texts and GPT-generated texts we used the Mann-Whitney U-test. This test was chosen for its suitability in comparing two independent samples, particularly since the sample sizes are small and not normally distributed. This test is useful for determining if there are differences in the median values between two sets of samples, and ranges from 0 to $N^2$ where $N$ is the number of samples in each set. In our case, the maximum possible U-test value is 100, since we have ten samples. Scores closer to 100 or 0 indicate a greater difference between sets, while scores closer to 50 indicate more overlap between the sets. We applied the Holm-Bonferroni correction for multiple comparisons.

\section{Results}
The results in Table \ref{tab:results} reveal significant differences between the citation structures in human-written and GPT-generated texts as exemplified by Table \ref{tab:samples}.
Example citation graphs and generated texts from the three conditions can be found in Table \ref{tab:graphs-abridged} and in Table \ref{tab:samples}, respectively. 
The full set of citation graphs can be found in Appendix \ref{app:graphs}. The text data can be viewed at \url{https://scholacite-data.streamlit.app/}.
Following are some key findings from our citation graph metrics:
\paragraph{Findings 1: Original human-written texts have significantly more edges:} The average number of edges was 48.3 in original texts, compared to only 12.3 for GPT-4 and 38.5 for GPT-assisted sections. This indicates that human authors integrate more citations into cohesive sentences that connect multiple relevant works. 

\begin{table*}[ht!]
    \centering
    \small
    \begin{tabular}{@{}p{0.31\textwidth}|p{0.31\textwidth}|p{0.31\textwidth}@{}}\toprule
        \textbf{Original} & \textbf{GPT-Assisted} & \textbf{GPT-4} \\
        \midrule
        \noindent
        \textbf{Knowledge Reasoning} Reasoning is the process of drawing new conclusions through the use of existing knowledge and rules. Progress has been reported in using PLMs to perform reasoning tasks, including arithmetic ([11]; [12]), commonsense ([13], [14]; [12]), logical ([15]) and symbolic reasoning ([12]). These abilities can be unlocked by finetuning a classifier on downstream datasets ([14]) or using proper prompting strategies (e.g., chain of thought (CoT) prompting ([12]) and generated knowledge prompting ([16])). This suggests that despite their insensitivity to negation ([17]; [18]) and over-sensitivity to lexicon cues like priming words ([19]; [20]), PLMs have the potential to make inferences over implicit knowledge and explicit natural language statements. In this work, we investigate the ability of PLMs to perform logical reasoning with implicit ontological knowledge to examine whether they understand the semantics beyond memorization. & 
        \noindent
        \textbf{Reasoning Abilities of PLMs} Recent research has closely examined the reasoning capacities of PLMs as they relate to ontological knowledge representation and processing ([6], [11], [12]). [6] analyze how PLMs break down complex inferential tasks into discrete logical steps, while [11] propose methods to strengthen contextual reasoning within model architectures. [12] find that step-by-step prompting enables PLMs to exhibit structured analytical thinking. These studies collectively spotlight sophisticated reasoning abilities within PLMs suitable for navigating ontological relationships. 

        \quad In other work, [14] test PLMs on drawing inferences from implicit world knowledge, with promising results, while [15] evaluate performance on formal logic challenges, revealing largely accurate deductive reasoning in areas involving ontological and commonsense reasoning. While displaying impressive reasoning skills, applying such capacities more broadly across knowledge domains remains an open challenge. Further inspection around the scope and limits of reasoning is warranted. & 
        \noindent
        The reasoning capabilities of PLMs are another central theme, with [6] examining large language models' ability to decompose high-level tasks into actionable steps. This concept of reasoning is further developed by [11], which proposes a new decoding strategy to improve the reasoning capabilities of PLMs. The paper by [12] aligns closely with our work by showing how chain-of-thought prompting can enhance reasoning in large language models. [14] demonstrates that PLMs can be trained to systematically reason over implicit knowledge, and [15] provides a comprehensive evaluation of logical reasoning tasks performed by these models. [19] rounds out this group with an overview of transformers' reasoning performance on various tasks, revealing the depth of their reasoning capabilities. \\
    \bottomrule
    \end{tabular}
    \caption{This table shows portions of the Related Work sections given the paper ``Do PLMs Know and Understand Ontological Knowledge?'' \cite{wu-etal-2023-plms}. The leftmost column contains the second paragraph of the original Related Work section. Since the Related Work sections generated by ScholaCite do not have a one-to-one alignment of citations with the original work, we selected portions of them that best overlap with the original paragraph in terms of topic (knowledge reasoning abilities of PLMs) and citations (citations 11, 12, 14, and 15 are discussed in all three samples and citation 19 is shared by the leftmost and rightmost samples). We can see that \textit{the original section demonstrates the most detailed knowledge of the cited works}. For example, the second sentence (beginning ``Progress has been reported...'') breaks down the papers about ``using PLMs to perform reasoning tasks'' according to the different types of reasoning (arithmetic, commonsense, logical, and symbolic). The purely GPT-generated text performs some synthesis of cited works, denoted by the phrases ``further developed by'', ``aligns closely with our work'', and ``rounds out this group,'' but does not compare works with the same detail. The GPT-assisted sample has a similar level of synthesis for this paper as GPT-4.}
    \label{tab:samples}
\end{table*}

\begin{table*}[]
    \centering
    \begin{tabular}{ccc}
        Human-Generated & AI-Assisted & GPT-Generated \\\midrule
            \includegraphics[width=0.3\textwidth]{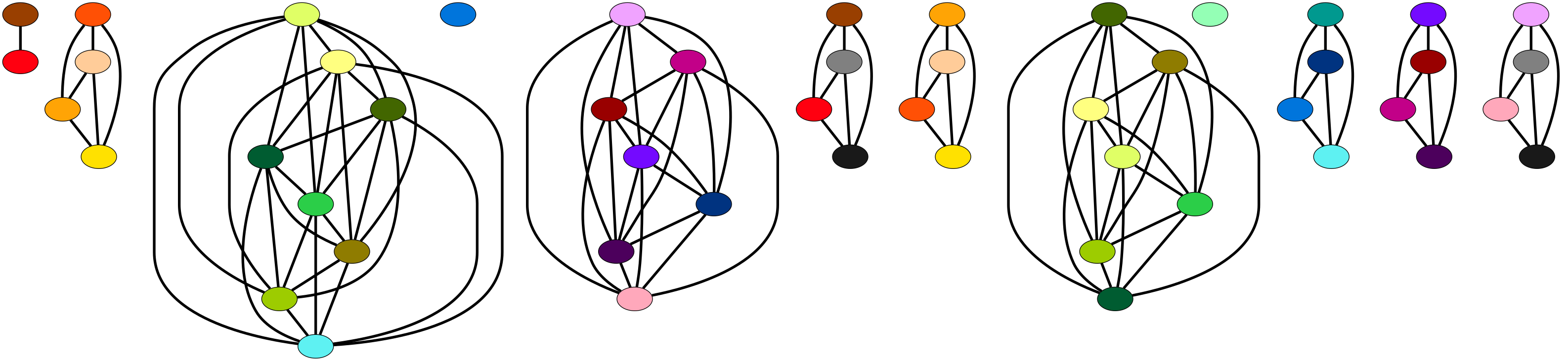}
             & \includegraphics[width=0.3\textwidth]{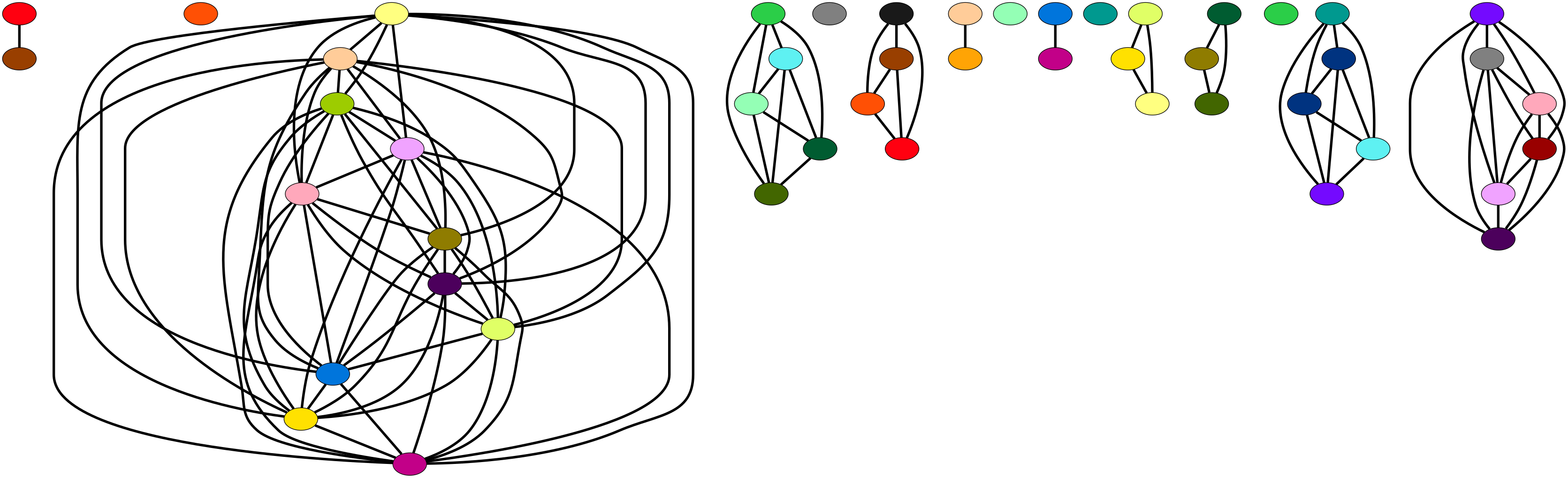} & \includegraphics[width=0.3\textwidth]{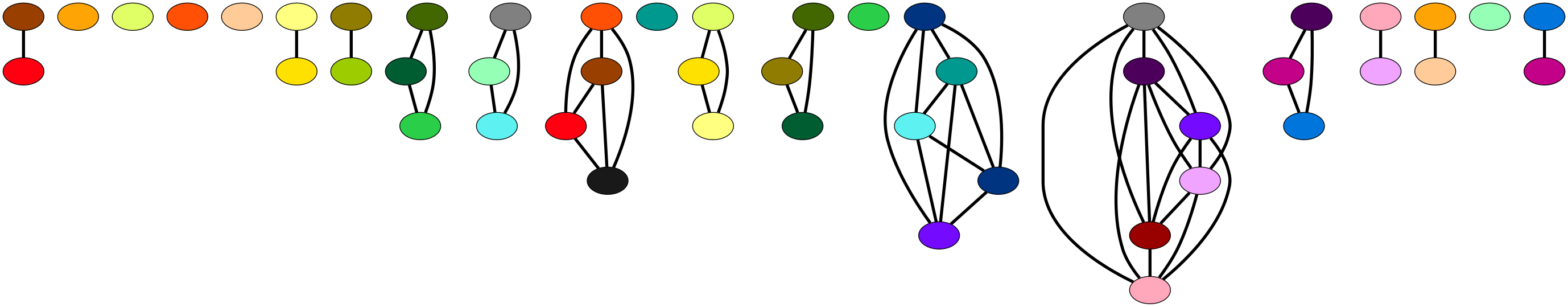}\\
             \midrule
             \includegraphics[width=0.24\textwidth]{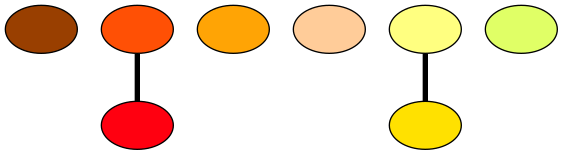}
             & \includegraphics[width=0.15\textwidth]{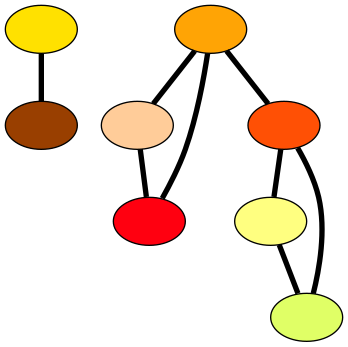} & \includegraphics[width=0.3\textwidth]{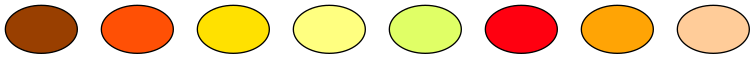}\\
             \midrule
             \includegraphics[width=0.25\textwidth]{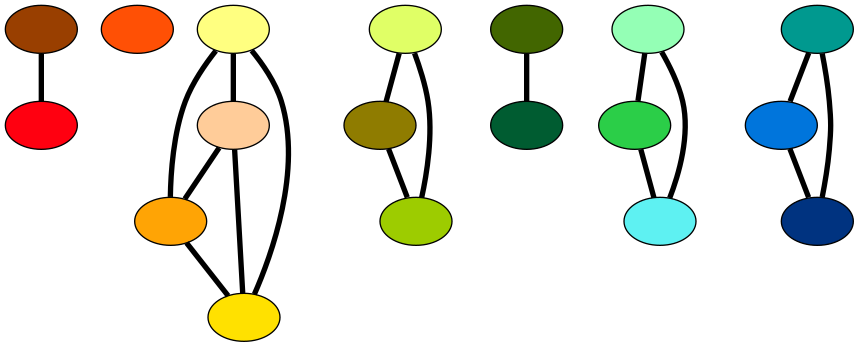}
             & \includegraphics[width=0.3\textwidth]{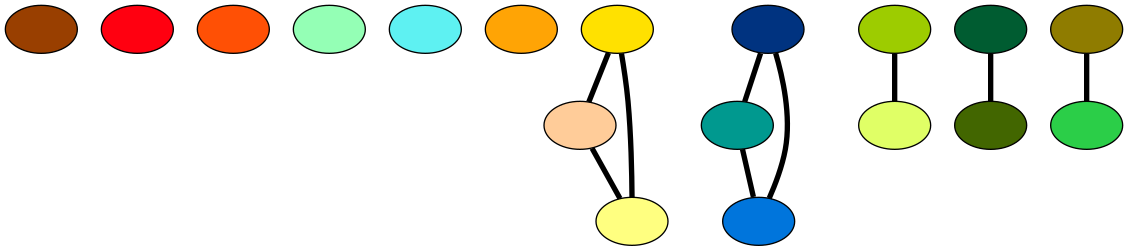} & \includegraphics[width=0.3\textwidth]{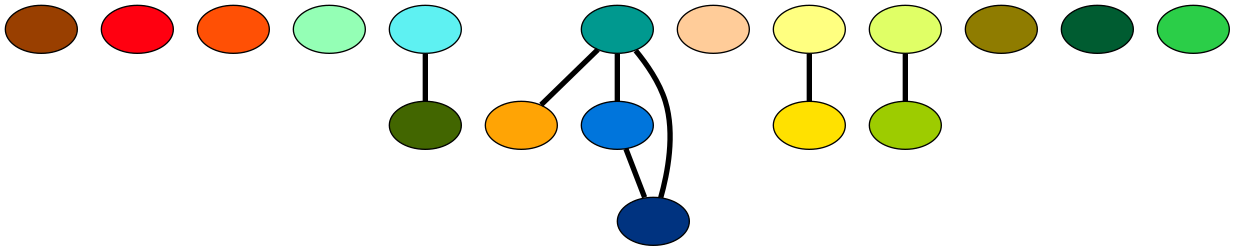}\\
             \bottomrule
         
    \end{tabular}
    \caption{Citation graphs for three of our samples. See Tables \ref{tab:graphs1} and \ref{tab:graphs2} in Appendix \ref{app:graphs} for the complete set.}
    \label{tab:graphs-abridged}
\end{table*}

\paragraph{Findings 2: Human-written and GPT-assisted texts have higher node degree:} Human-written texts had a higher average node degree of 3.23 compared to 0.7 for GPT-generated and 2.69 for GPT-assisted texts. This shows that in human-written and GPT-assisted texts, individual citations were more frequently connected to multiple other works rather than discussed in isolation. 

\paragraph{Findings 3: Density is significantly higher for human-written and GPT-assisted conditions:} While all texts had an overall sparse citation network structure, the density was significantly higher in original texts (0.14) and GPT-assisted (0.13) compared to GPT-generated (0.02). This reinforces the finding that the human-written texts have a more interconnected citation structure. 

\paragraph{Findings 4: Greater clustering in human-written and GPT-assisted texts:} The clustering coefficient indicates subgrouping of citations. Original human-written texts show an average clustering coefficient of 0.66 and GPT-assisted was 0.63, significantly higher than 0.15 for GPT. This difference suggests that the human-written and collaborative sections tend to cluster citations in better-defined subgroups, possibly indicating more cohesive groupings of related works.

These results show a disparity in the integration and interconnectedness of citations between human-written and GPT-generated texts. The original sections show a more detailed engagement with existing literature, containing more complex networks of citations. This complexity, represented by higher edge counts, node degrees, density, and clustering coefficients, show limitations of GPT-4 in replicating the more sophisticated discourse found in academic writing. As we move forward as a research community, we should focus on GPT-assisted writing tools that focus on repetitive and less intellectually intense aspects of academic writing.

\section{Conclusion}
Using citation graph metrics, we presented an analysis comparing human-written, GPT-assisted, and GPT-generated related work sections. Our goal was to evaluate the ability of LLM text to perform the synthesis and framing expected in quality academic writing. We find that without human intervention, GPT-4 fails at interconnecting citations to the degree seen in human-written sections. This suggests fundamental limitations of the model in achieving deep scholarly knowledge integration from only abstract-level data. LLMs operate on a limited context window, restricting their ability to connect concepts across very longs texts. Fine-tuning specifically for synthesis tasks could improve performance. However, with appropriate and detailed prompting and scholar oversight, LLMs may be able to play a beneficial assisting role.

While purely GPT-generated text shows a lack of citation cohesion, through our ScholaCite system we demonstrate a collaborative approach that yields citation networks indistinguishable from human expert writing. This points to the importance of human guidance in applying large language models, rather than fully automating writing tasks. Our analysis provides both cautionary notes regarding over-reliance on AI and promising indications that LLMs can be effectively integrated into the research process when thoughtfully implemented under domain expert supervision. We recommend that AI-assisted writing tools should avoid drafting full sections of text to avoid simplified and generic discourse.

\section{Ethical Considerations}
The rapid advancement of large language models demands careful ethical consideration regarding their integration into academic writing workflows. As demonstrated through our analysis, fully automated scholarly writing risks severely compromising standards of knowledge synthesis and citation integration. When employing such models for academic purposes, it is important to be aware of the potential risks including propagated inaccuracies and plagiarism. Proactive measures such as transparency in LLM usage, implementing copyright protections, verification processes, and emphasizing semantically-meaningful prompting will be important for addressing these concerns.

\section{Limitations}
Our sample size of 10 papers necessarily restricts the generalizability and statistical power of our work. Additionally, examining only best papers from the 2023 ACL conference proceedings does not capture the full diversity of the NLP domain. Follow-up studies should incorporate other fields and venue types. There is also an inherent subjectivity in determining ``high-quality'' related work sections for comparison. Our evaluation based on citation graph analysis, while allowing scalable automated analysis, does not measure all facets that constitute quality, but rather should be seen as a proxy measure for cited work synthesis. Further layers of qualitative assessment could reveal additional capabilities and deficiencies of LLM-generated text. Our paper explores only one set of hyperparameters and prompt design choices. Varying factors like context window size and fine-tuning approaches could potentially improve synthesis capabilities. However, even highly interconnected citation graphs may lack deeper integration of concepts. Our graph-based metric serves as an initial indicator of quality, but has inherent limitations in fully evaluating the nuance of scholarly discourse. Finally, our paper does not contain an evaluation of the ScholaCite tool itself, which was used by only one author of this paper. We save a human study for a future paper in the Human-Computer Interaction domain.

\FloatBarrier

\bibliography{main}

\appendix
\section{GPT-4 Prompts}\label{app:prompts}
\subsection{Prompts for Step 1: Generation of Citation Groups}
\subsubsection{Step 1a}
I will provide you with the abstract, introduction, and conclusion sections of an in-progress academic paper, along with a list of relevant scholarly articles. Each article is identified by a unique ID. Your task is to organize these articles into thematic citation groups that align with and support the framing of the in-progress paper. For each citation group, you should: 
\begin{enumerate}
    \item Name the Group: Assign a descriptive name that captures the thematic focus of the group; 
    \item Explain the Rationale: Detail how the articles within this group contribute to or support the thematic framework of the in-progress paper; and
    \item List the Cited Papers: Include the articles that fall under this thematic group, providing for each article the citation ID and the title. 
\end{enumerate}

It's important that each article is placed in at least one group, although articles may fit into multiple groups if applicable. 

Your response should be formatted as a JSON object, where each key represents a unique group index. The value for each key should be a dictionary with three main keys: `group\_name', `group\_rationale', and `cited\_papers'. The `cited\_papers' should be a list of dictionaries, each containing `id' and `title' keys for the articles. Here is the structure for your output:
\begin{verbatim}
{
  "1": {
    "group_name": "Name of first group", 
    "group_rationale": "This rationale 
                       explains why this 
                       group should exist 
                       and how it helps 
                       frame the 
                       work-in-progress.", 
    "cited_papers": [
      {
        "id": "1", 
        "title": "Title of the first paper 
                 included in the group"
      }, 
      {
        "id": "3", 
        "title": "Title of the second paper 
                 included in group 1"
      }
    ]
  }, 
  "2": {
    "group_name": "Name of second group", 
    "group_rationale": "This rationale 
                       explains why this 
                       group should exist 
                       and how it helps 
                       frame the 
                       work-in-progress.", 
    "cited_papers": [
      {
        "id": "2", 
        "title": "Title of the first paper 
                 included in the group"
      }, 
      {
        "id": "3", 
        "title": "Title of the second paper 
                 included in group 2"
      }
    ]
  }
}
\end{verbatim}
Title: \emph{\textcolor{red}{title of the in-progress paper}}

\noindent Abstract: \emph{\textcolor{red}{abstract of the in-progress paper}}

\noindent Introduction

\noindent\emph{\textcolor{red}{introduction of the in-progress paper}}

\noindent Conclusion

\noindent\emph{\textcolor{red}{conclusion of the in-progress paper}}

\vspace{0.5cm}
\noindent Related Works:

\noindent 1. \emph{\textcolor{red}{title of citation 1}}

\vspace{0.5cm}
\noindent 2. \emph{\textcolor{red}{title of citation 2}}

\ \ \ \ \ \ \ \ \ \ .

\ \ \ \ \ \ \ \ \ \ .

\ \ \ \ \ \ \ \ \ \ .

\subsubsection{Step 1b}
I will provide you with the abstract, introduction, and conclusion sections of an in-progress academic paper, a data structure showing a citation group and the titles of the papers assigned to the group, and the abstracts of each of the cited works and their IDs. I need you to update the information in the data structure by doing the following:
\begin{enumerate}
    \item provide rationales for the inclusion of each work in its group based on the provided abstract;
    \item  provide the span of text from the abstract that supports your rationale; and
    \item output the revised data structure following a pattern I will give you.
\end{enumerate}
The input data is a dictionary where each key is a group index. Each value is a dictionary with three top-level keys: `group\_name', `group\_rationale', and `cited\_papers'. The value for papers is a list of dictionaries. Each dictionary has two keys: `id', and `title'. Here is the structure for the input:
\begin{verbatim}
{
  "1": {
    "group_name": "Name of first group", 
    "group_rationale": "This rationale 
                       explains why this 
                       group should exist 
                       and how it helps 
                       frame the 
                       work-in-progress.", 
    "cited_papers": [
      {
        "id": "1", 
        "title": "Title of the first paper 
                 included in the group"
      }, 
      {
        "id": "3", 
        "title": "Title of the second paper 
                 included in the group"
      }
    ]
  }
}
\end{verbatim}

The output data will have the same structure, except that the data will be updated for each cited work to include the following keys: `id', `title', `citation\_rationale', and `span'. Here is the structure for your output:
\begin{verbatim}
{
  "1": {
    "group_name": "Name of first group", 
    "group_rationale": "This rationale 
                       explains why this 
                       group should exist 
                       and how it helps 
                       frame the 
                       work-in-progress.", 
    "cited_papers": [
      {
        "id": "1", 
        "title": "Title of the first paper 
                 included in the group", 
        "citation_rationale": "the rationale 
                              for this paper's 
                              inclusion in the 
                              group", 
        "span": "The span of text from the 
                abstract that supports the 
                rationale"
      }, 
      {
        "id": "3", 
        "title": "Title of the second paper 
                 included in the group", 
        "citation_rationale": "the rationale 
                              for this paper's 
                              inclusion in the 
                              group", 
        "span": "The span of text from the 
                abstract that supports the 
                rationale"
      }
    ]
  }
}
\end{verbatim}
Please ensure your response consists solely of the fully formed JSON data structure, with no additional text outside of the JSON formatting. Ensure no entry from the JSON data structure is omitted.

\vspace{0.5cm}
\noindent Work-in-progress data:

\noindent Title: \emph{\textcolor{red}{title of the in-progress paper}}

\noindent Abstract: \emph{\textcolor{red}{abstract of the in-progress paper}}

\noindent Introduction

\noindent\emph{\textcolor{red}{introduction of the in-progress paper}}

\noindent Conclusion

\noindent\emph{\textcolor{red}{conclusion of the in-progress paper}}

\vspace{0.5cm}
\noindent Related Works:

\noindent ID: 1

\noindent title: \emph{\textcolor{red}{title of citation 1}}

\noindent abstract: \emph{\textcolor{red}{abstract of citation 1}}

\vspace{0.5cm}
\noindent ID: 2

\noindent title: \emph{\textcolor{red}{title of citation 2}}

\noindent abstract: \emph{\textcolor{red}{abstract of citation 2}}

\ \ \ \ \ \ \ \ \ \ .

\ \ \ \ \ \ \ \ \ \ .

\ \ \ \ \ \ \ \ \ \ .

\subsection{Prompt for Step 2: Generation of Related Work Section}
Given the context of an in-progress scholarly paper and a set of groupings of related works, your task is to generate the related work section of the work-in-progress. The groupings will be provided to you in json format. The top-level keys are numbers, which are the group indices. Each value is another dictionary, which has the following keys: 
\begin{enumerate}
    \item `group\_name' (the name of the group);
    \item `group\_rationale' (an explanation of how the group supports and frames the scholarly work-in-progress); and
    \item `cited\_papers' (the list of papers that belong to the group).
\end{enumerate}
Each entry in the cited\_papers list is another dictionary, which has the following keys: 
\begin{enumerate}
    \item `id' (the citation ID);
    \item `title': (the title of the citation);
    \item `citation\_rationale' (an explanation of why this cited work should be in this citation group); and
    \item `span' (the span of text from the work's abstract that supports the citation\_rationale).
\end{enumerate} Cite the papers using the citation ID. Structure the text so that each grouping is a distinct paragraph. Do not reference the fact that the text was formed from groups (e.g. do not say ``The group `Data Mining Approaches' includes citations [1] through [2]''). 

\vspace{0.5cm}
\noindent Groupings:

\noindent\emph{\color{red}{Groupings data structure here}}

\vspace{0.5cm}
\noindent Work-in-progress data:

\noindent Title: \emph{\textcolor{red}{title of the in-progress paper}}

\noindent Abstract: \emph{\textcolor{red}{abstract of the in-progress paper}}

\noindent Introduction

\noindent\emph{\textcolor{red}{introduction of the in-progress paper}}

\noindent Conclusion

\noindent\emph{\textcolor{red}{conclusion of the in-progress paper}}

\section{Graph Representations}\label{app:graphs}
See Tables \ref{tab:graphs1} and \ref{tab:graphs2} for the citation networks for each data sample.

\begin{table*}[]
    \centering
    \begin{tabular}{ccc}
        Human-Generated & AI-Assisted & GPT-Generated \\\midrule
            \includegraphics[width=0.3\textwidth]{Figures/graphs/31-sentence-ground.png}
             & \includegraphics[width=0.3\textwidth]{Figures/graphs/31-sentence-colab.png} & \includegraphics[width=0.3\textwidth]{Figures/graphs/31-sentence-gpt.png}\\
             \multicolumn{3}{p{0.9\textwidth}}{``World-to-Words: Grounded Open Vocabulary Acquisition through Fast Mapping in Vision-Language Models'' \cite{ma-etal-2023-world}.} \\\midrule
             \includegraphics[width=0.24\textwidth]{Figures/graphs/36-sentence-ground.png}
             & \includegraphics[width=0.14\textwidth]{Figures/graphs/36-sentence-colab.png} & \includegraphics[width=0.3\textwidth]{Figures/graphs/36-sentence-gpt.png}\\
             \multicolumn{3}{p{0.9\textwidth}}{``When Does Translation Require Context? A Data-driven, Multilingual Exploration'' \cite{fernandes-etal-2023-translation}.} \\\midrule
             \includegraphics[width=0.23\textwidth]{Figures/graphs/38-sentence-ground.png}
             & \includegraphics[width=0.3\textwidth]{Figures/graphs/38-sentence-colab.png} & \includegraphics[width=0.3\textwidth]{Figures/graphs/38-sentence-gpt.png}\\
             \multicolumn{3}{p{0.9\textwidth}}{``LexSym: Compositionality as Lexical Symmetry'' \cite{akyurek-andreas-2023-lexsym}.} \\\midrule
             \includegraphics[width=0.3\textwidth]{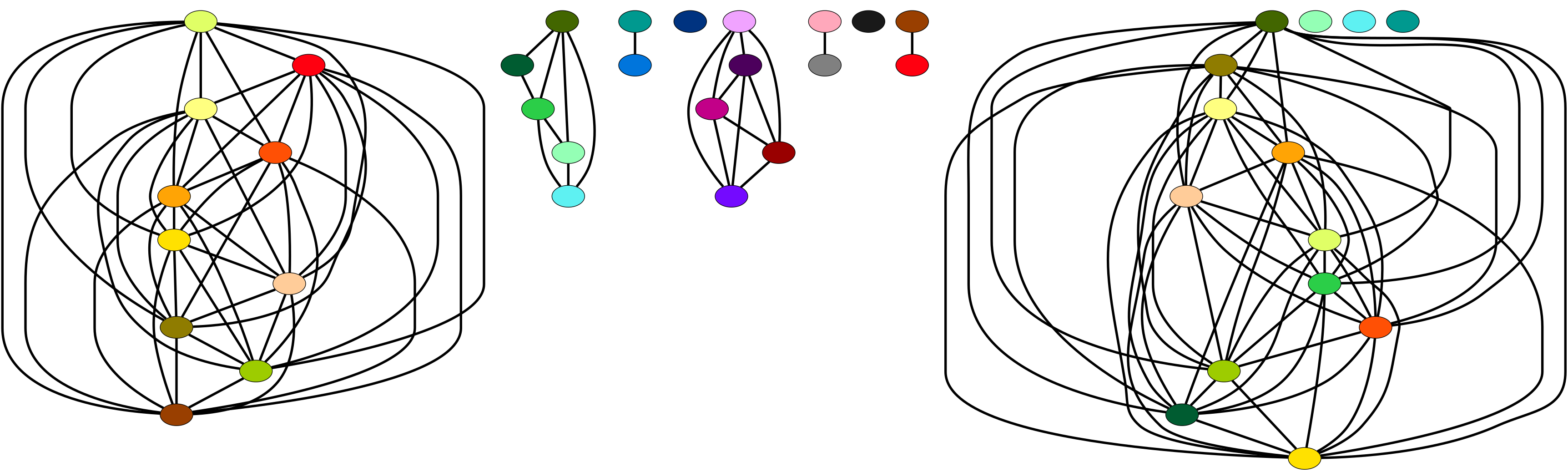}
             & \includegraphics[width=0.3\textwidth]{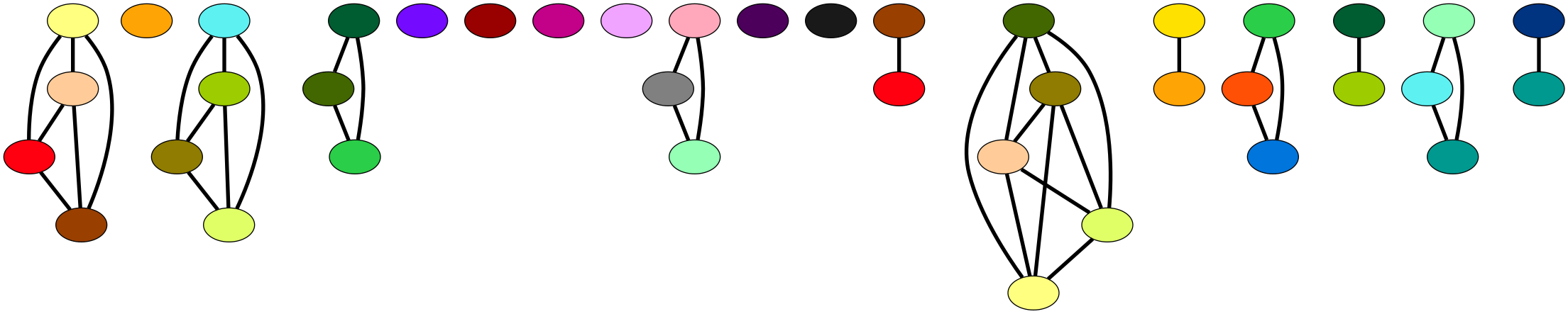} & \includegraphics[width=0.3\textwidth]{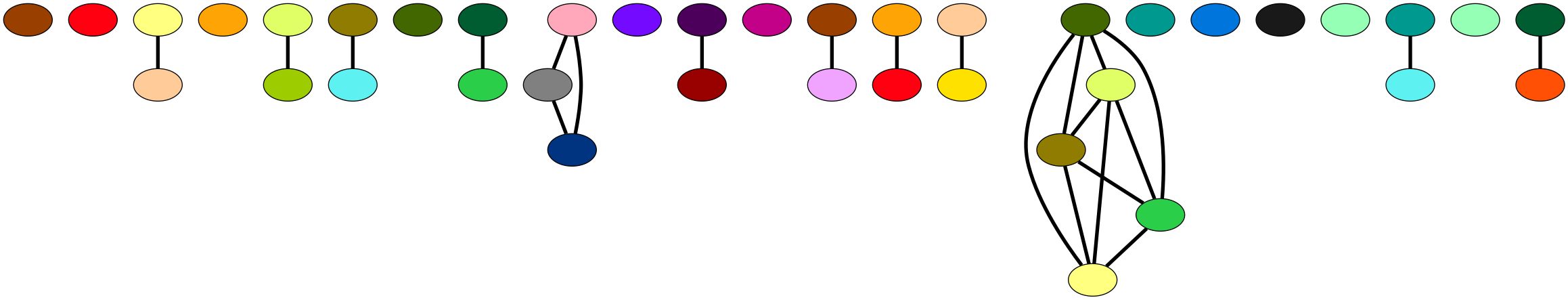}\\
             \multicolumn{3}{p{0.9\textwidth}}{``Do Androids Laugh at Electric Sheep? Humor “Understanding” Benchmarks from The New Yorker Caption Contest'' \cite{hessel-etal-2023-androids}.} \\\midrule
             \includegraphics[width=0.3\textwidth]{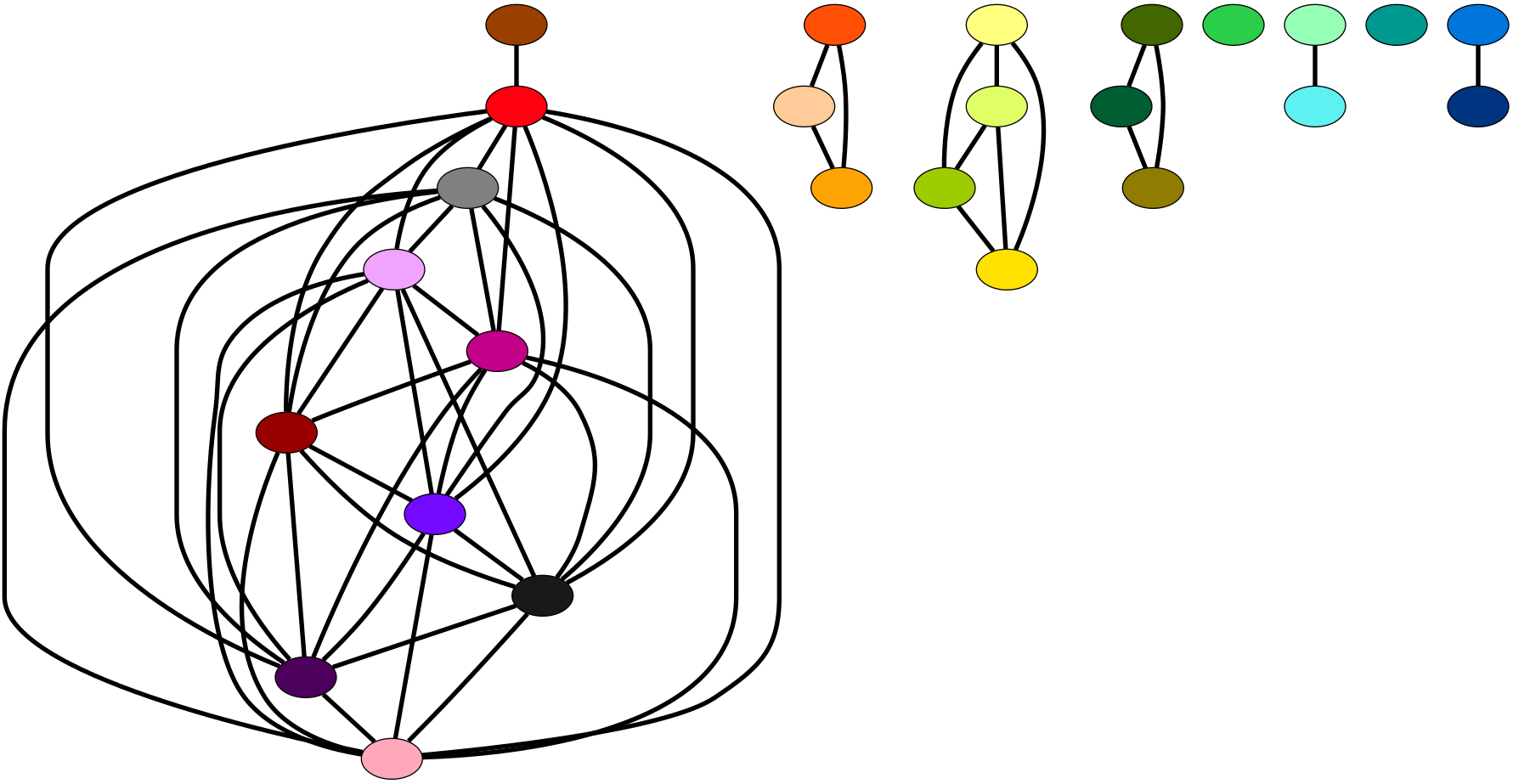}
             & \includegraphics[width=0.25\textwidth,trim={0 0 0 3.5cm},clip]{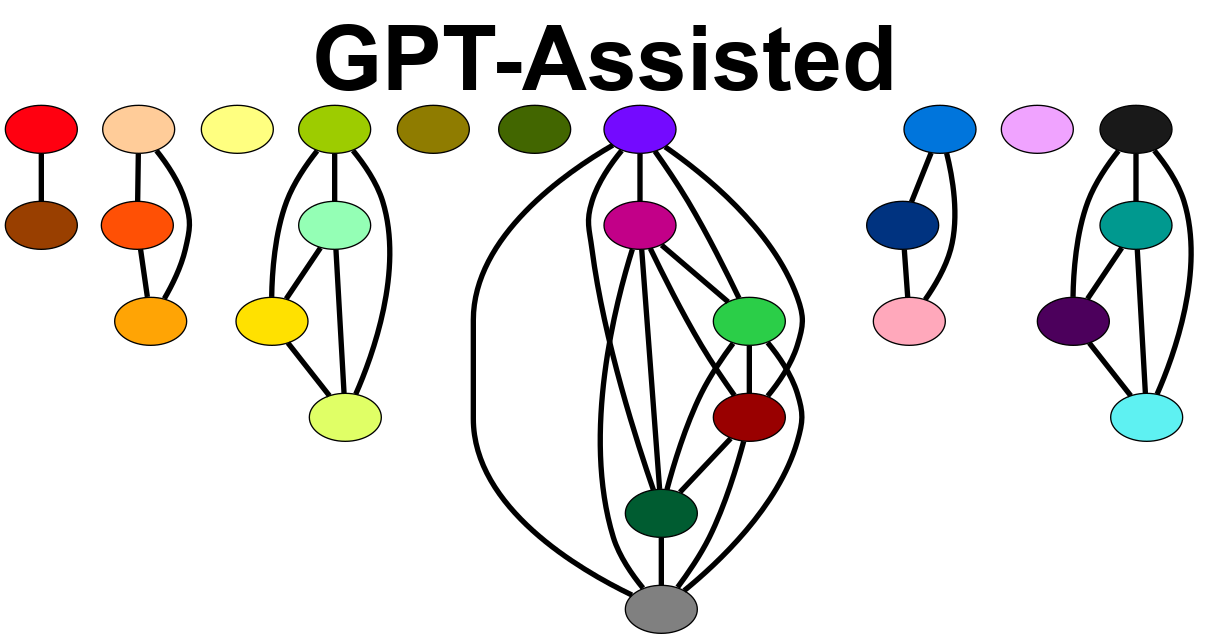} & \includegraphics[width=0.3\textwidth]{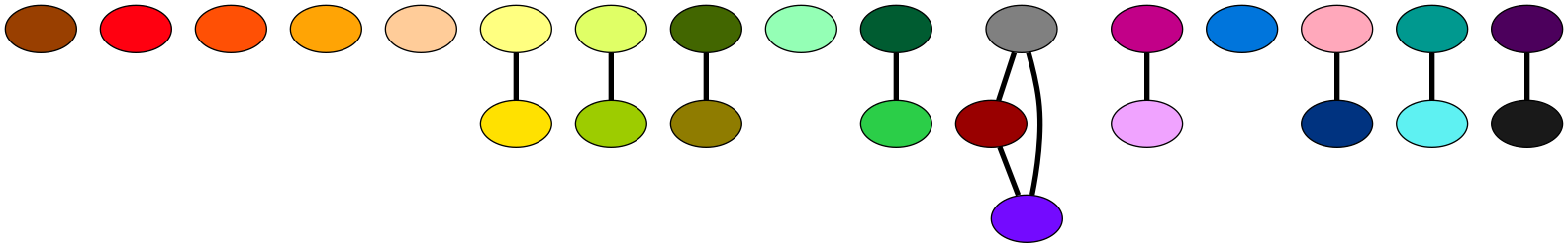}\\
             \multicolumn{3}{p{0.9\textwidth}}{``Glot500: Scaling Multilingual Corpora and Language Models to 500 Languages'' \cite{imanigooghari-etal-2023-glot500}.} \\\bottomrule
        \end{tabular}
    \caption{Citation graphs for samples 1-5.}
    \label{tab:graphs1}
\end{table*}

\begin{table*}[]
    \centering
    \begin{tabular}{ccc}
        Human-Generated & AI-Assisted & GPT-Generated \\\midrule
             \includegraphics[width=0.2\textwidth]{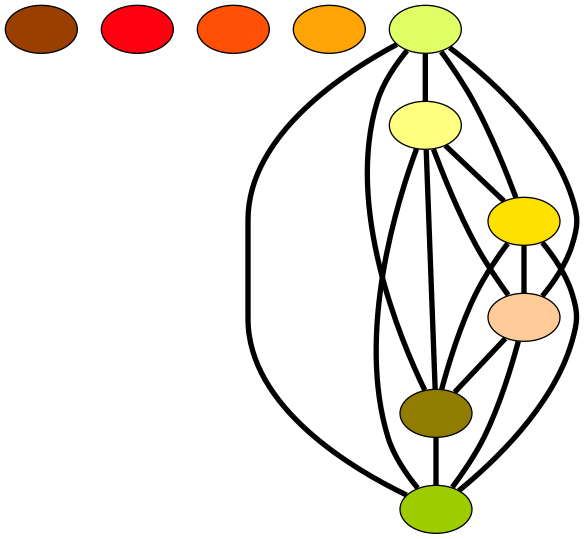}
             & \includegraphics[width=0.17\textwidth]{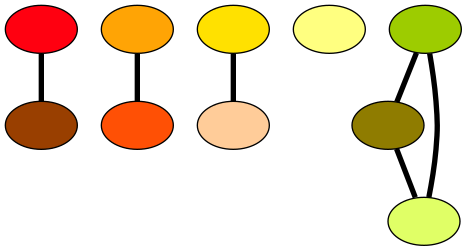} & \includegraphics[width=0.3\textwidth]{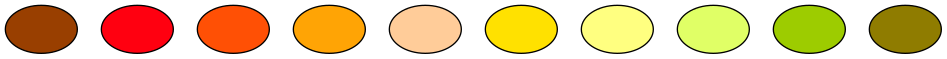}\\
             \multicolumn{3}{p{0.9\textwidth}}{``Improving Pretraining Techniques for Code-Switched NLP'' \cite{das-etal-2023-improving}.} \\\midrule
             \includegraphics[width=0.3\textwidth]{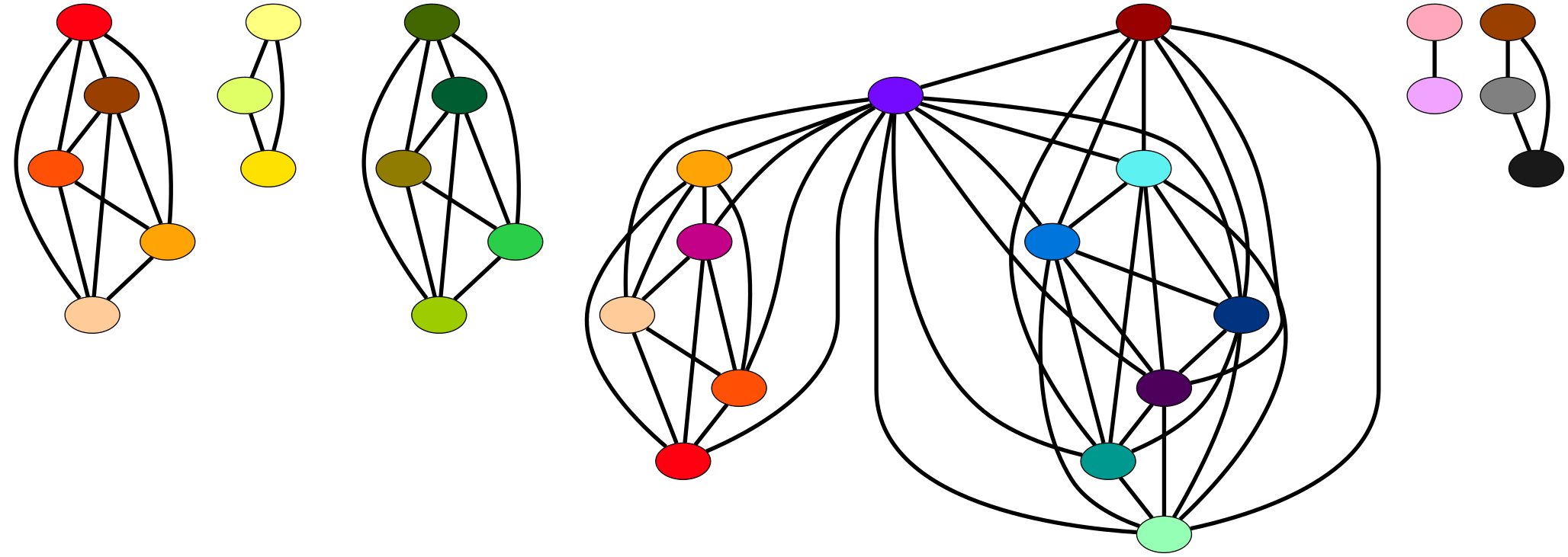}
             & \includegraphics[width=0.24\textwidth]{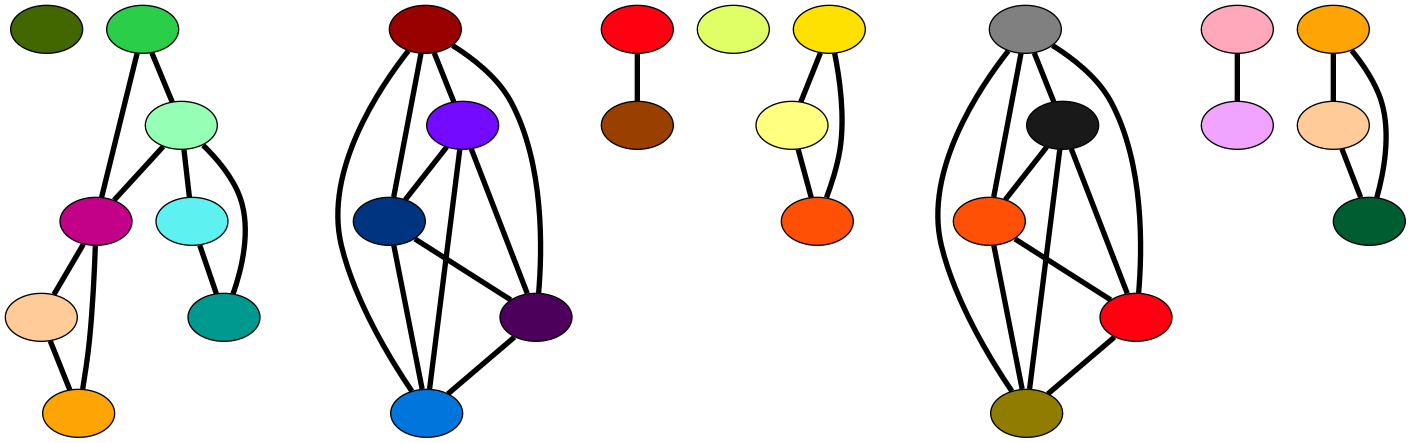} & \includegraphics[width=0.3\textwidth]{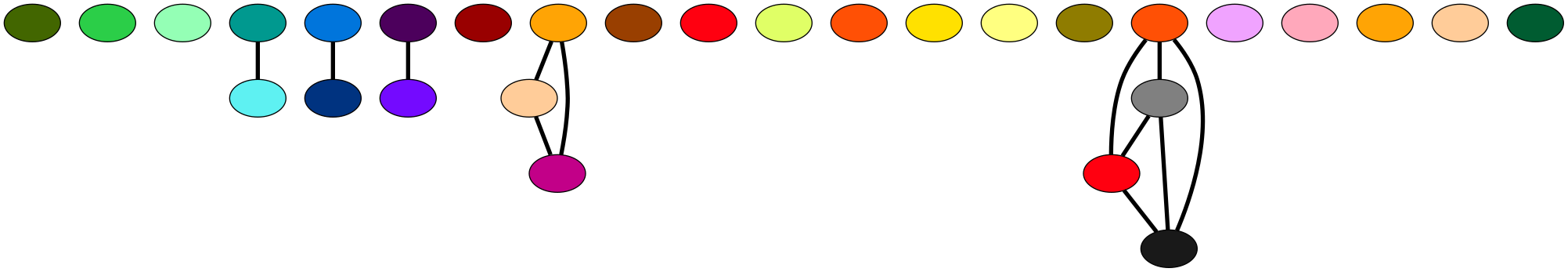}\\
             \multicolumn{3}{p{0.9\textwidth}}{``Marked Personas: Using Natural Language Prompts to Measure Stereotypes in Language Models'' \cite{cheng-etal-2023-marked}.} \\\midrule
             \includegraphics[width=0.25\textwidth]{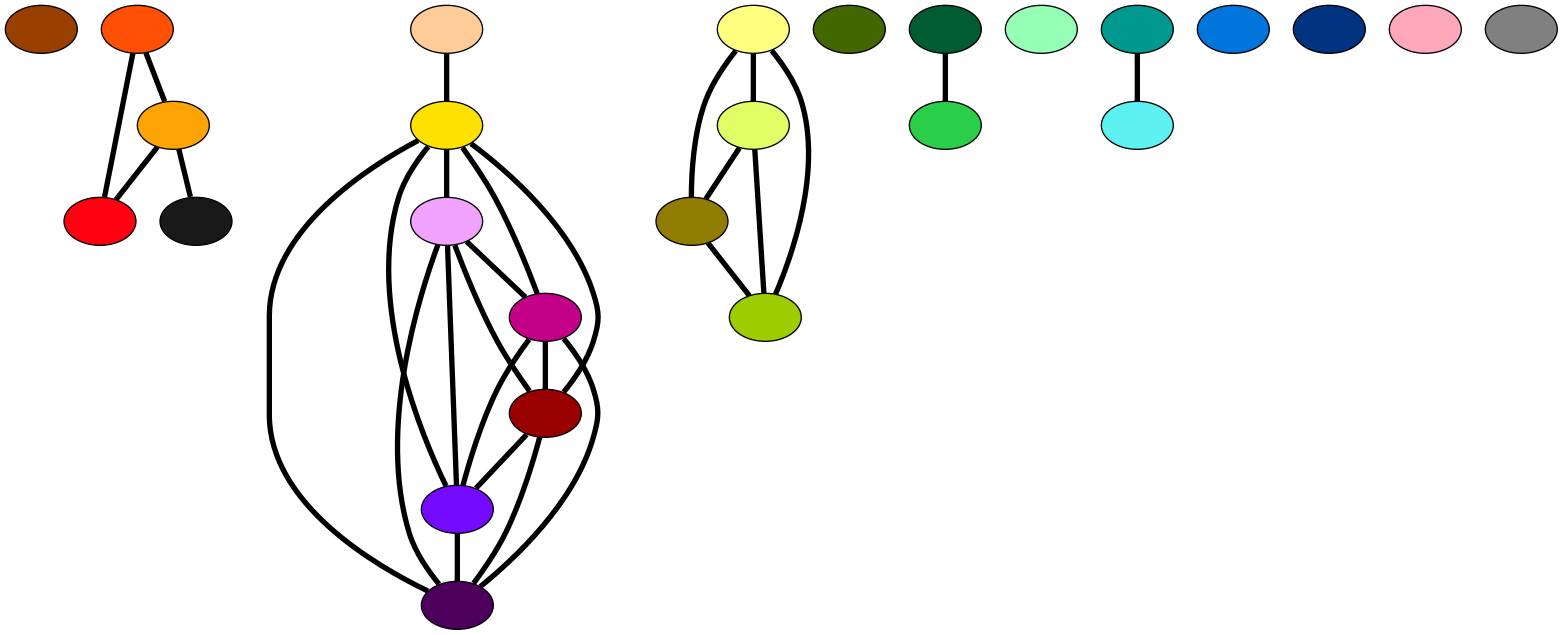}
             & \includegraphics[width=0.17\textwidth]{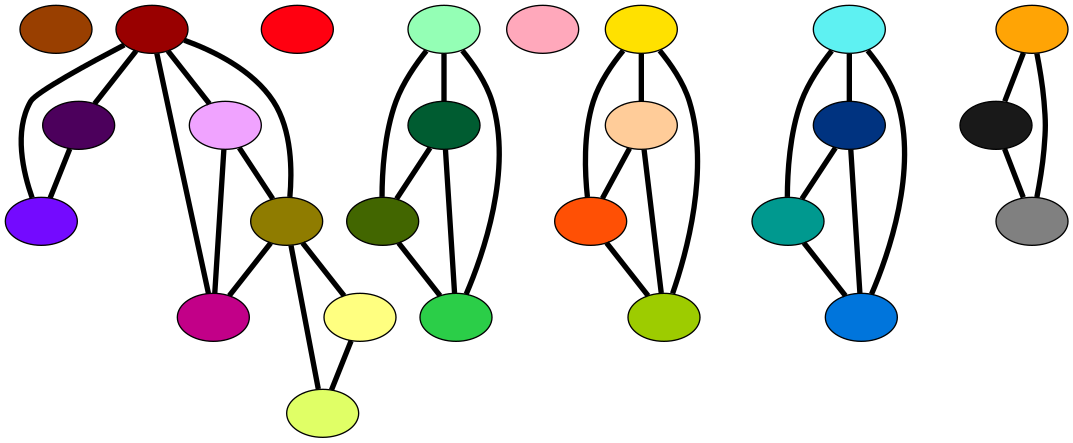} & \includegraphics[width=0.3\textwidth]{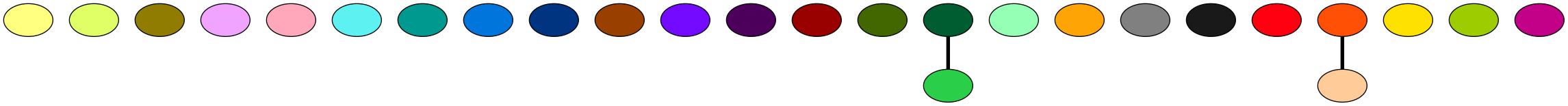}\\
             \multicolumn{3}{p{0.9\textwidth}}{``Small Data, Big Impact: Leveraging Minimal Data for Effective Machine Translation'' \cite{maillard-etal-2023-small}.} \\\midrule
             \includegraphics[width=0.3\textwidth]{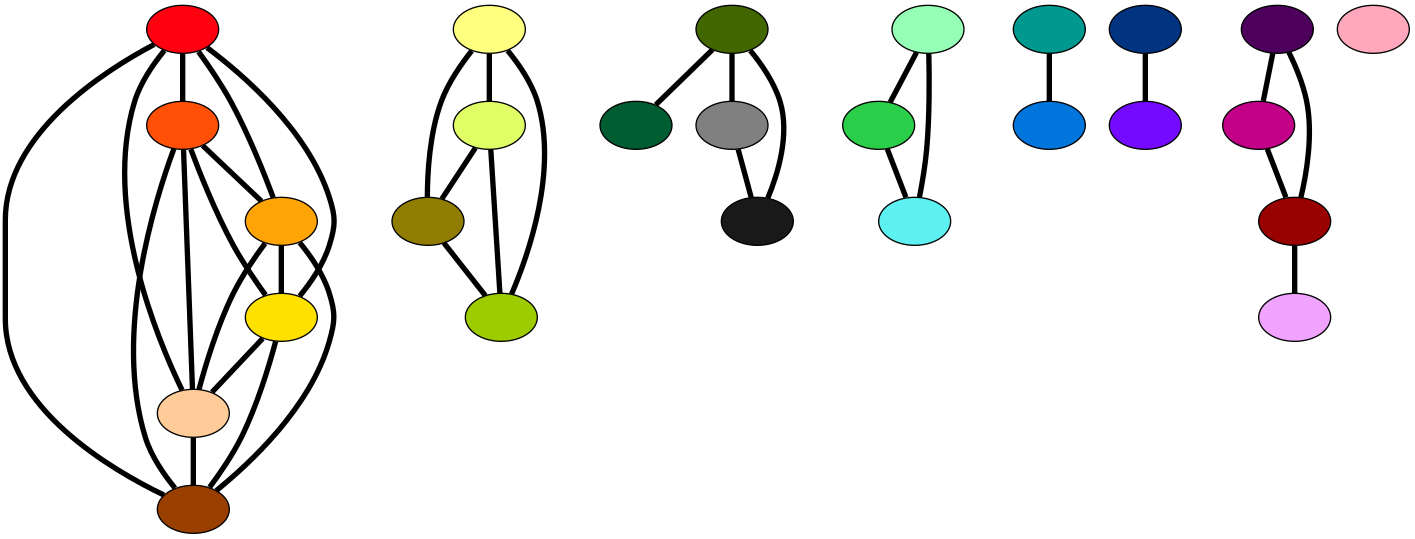}
             & \includegraphics[width=0.3\textwidth]{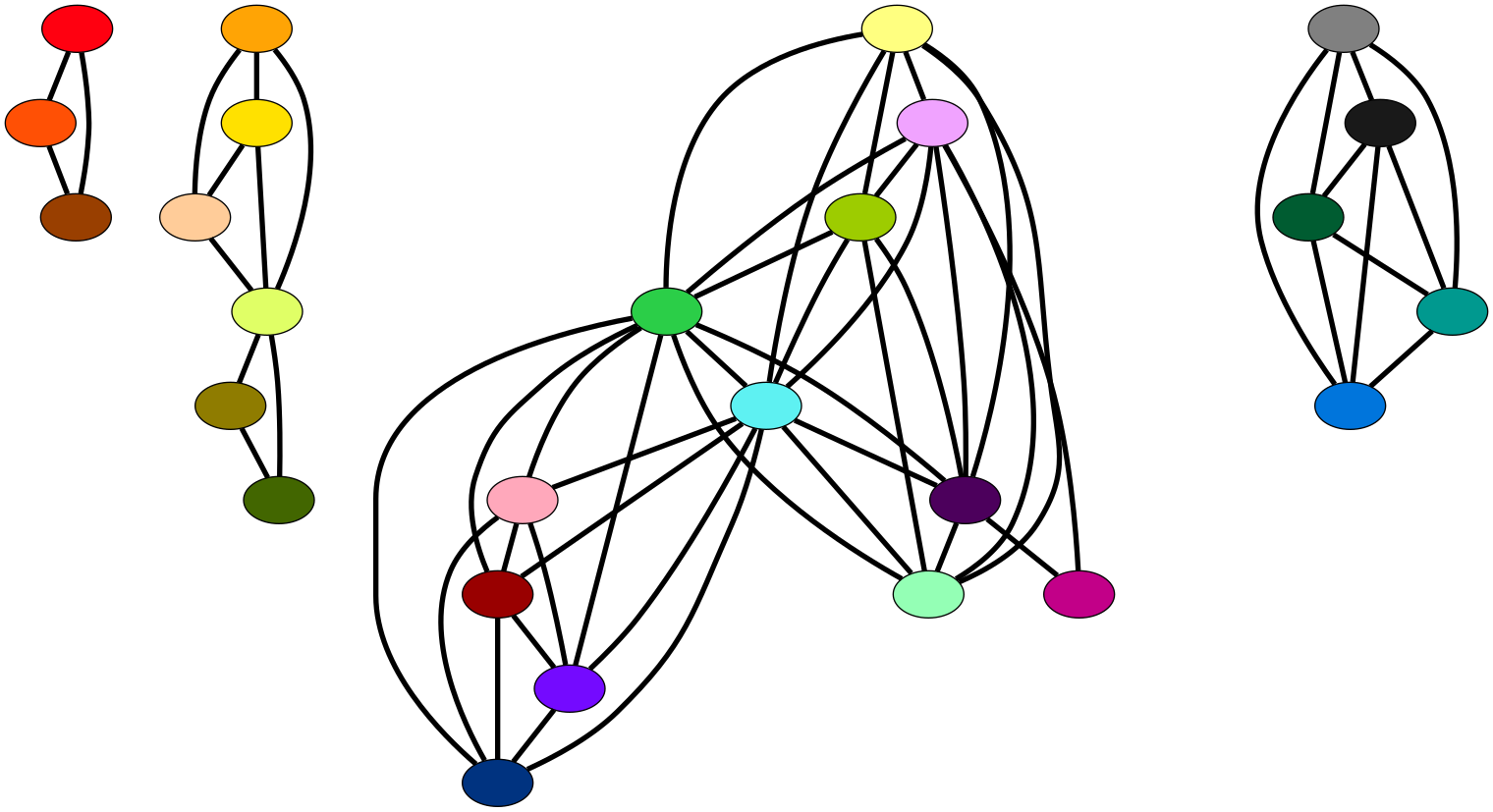} & \includegraphics[width=0.3\textwidth]{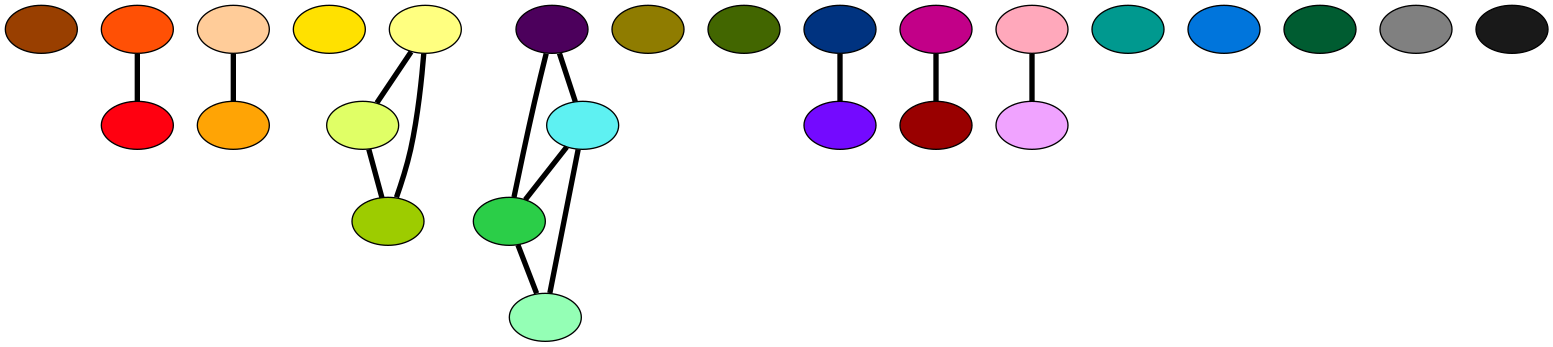}\\
             \multicolumn{3}{p{0.9\textwidth}}{``Faithful Low-Resource Data-to-Text Generation through Cycle Training''\cite{wang-etal-2023-faithful}.} \\\midrule
             \includegraphics[width=0.18\textwidth]{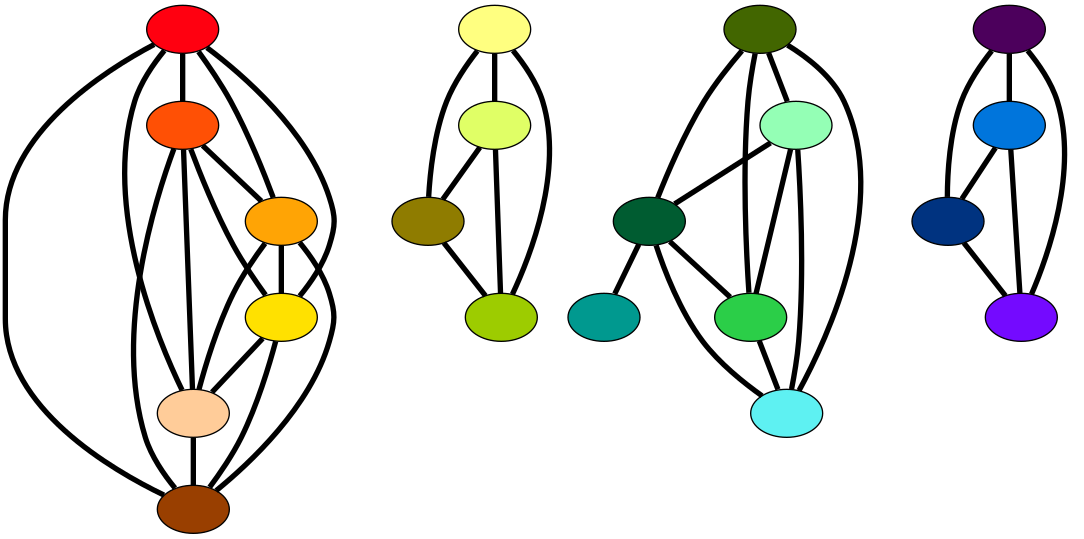}
             & \includegraphics[width=0.18\textwidth]{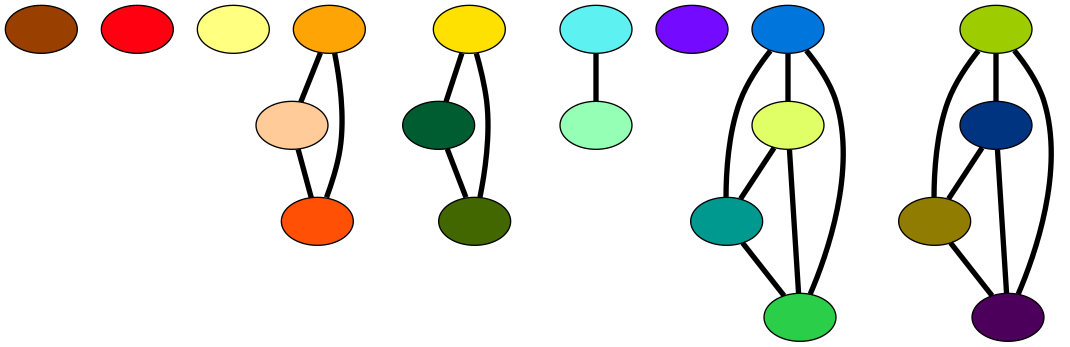} & \includegraphics[width=0.3\textwidth]{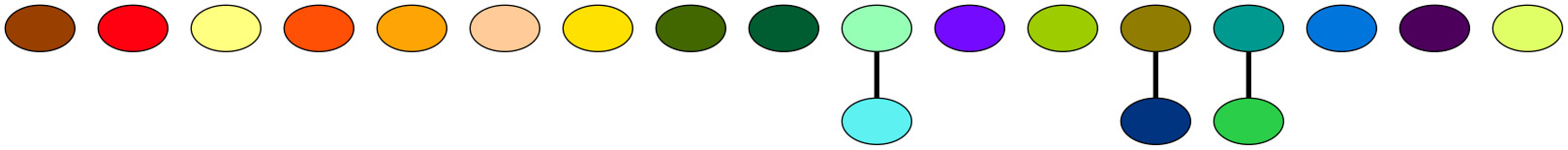}\\
             \multicolumn{3}{p{0.9\textwidth}}{``Do PLMs Know and Understand Ontological Knowledge?''\cite{wu-etal-2023-plms}.} \\\bottomrule
         
    \end{tabular}
    \caption{Citation graphs for samples 6-10.}
    \label{tab:graphs2}
\end{table*}

\end{document}